\documentclass{article} 
\usepackage[preprint]{colm2026_conference}

\usepackage{microtype}
\usepackage{hyperref}
\usepackage{url}
\usepackage{booktabs}
\usepackage{graphicx}
\usepackage{amsmath}
\usepackage{amsfonts}
\usepackage{algorithm}
\usepackage{algorithmicx}
\usepackage{algcompatible}
\usepackage{subcaption}
\usepackage[most]{tcolorbox}
\usepackage{xcolor}
\usepackage{verbatim}
\usepackage{fvextra}
\usepackage{pgfplots}
\usepackage{multirow}
\usepackage{placeins}

\newtcolorbox{promptbox}[1]{
    colback=gray!5,
    colframe=black!75,  
    fonttitle=\bfseries,
    coltitle=white,     
    title={#1},          
    enhanced,          
    title={#1},
    arc=3pt,            
    boxrule=0.8pt,      
    left=5pt, right=5pt, top=10pt, bottom=5pt, 
    fontupper=\small,   
    breakable
}

\usepackage{lineno}

\definecolor{darkblue}{rgb}{0, 0, 0.5}
\hypersetup{colorlinks=true, citecolor=darkblue, linkcolor=darkblue, urlcolor=darkblue}

\title{AIVV: Neuro-Symbolic LLM Agent-Integrated Verification and Validation for Trustworthy Autonomous Systems}

\author{
  Jiyong Kwon$^{1}$ \quad
  Ujin Jeon$^{2}$ \quad
  Sooji Lee$^{3}$ \quad
  Guang Lin$^{1,4}$ \\
  \\
  $^{1}$School of Mechanical Engineering, Purdue University \\
  $^{2}$School of Electrical and Computer Engineering, Purdue University \\
  $^{3}$Department of Computer Science, Purdue University \\
  $^{4}$Department of Mathematics, Purdue University \\
  West Lafayette, IN 47907, USA \\
  \texttt{\{kwon165, ujeon, lee5264, guanglin\}@purdue.edu}
}

\begin{document}

\ifcolmsubmission
\linenumbers
\fi

\maketitle



\begin{abstract}
Deep learning models excel at detecting anomaly patterns in normal data. However, they do not provide a direct solution for anomaly classification and scalability across diverse control systems, frequently failing to distinguish genuine faults from nuisance faults caused by noise or the control system's large transient response. Consequently, because algorithmic fault validation remains unscalable, full Verification and Validation (V\&V) operations are still managed by Human-in-the-Loop (HITL) analysis, resulting in an unsustainable manual workload. To automate this essential oversight, we propose Agent-Integrated Verification and Validation (AIVV), a hybrid framework that deploys Large Language Models (LLMs) as a deliberative outer loop. Because rigorous system verification strictly depends on accurate validation, AIVV escalates mathematically flagged anomalies to a role-specialized LLM council. The council agents perform collaborative validation by semantically validating nuisance and true failures based on natural-language (NL) requirements to secure a high-fidelity system-verification baseline. Building on this foundation, the council then performs system verification by assessing post-fault responses against NL operational tolerances, ultimately generating actionable V\&V artifacts, such as gain-tuning proposals. Experiments on a time-series simulator for Unmanned Underwater Vehicles (UUVs) demonstrate that AIVV successfully digitizes the HITL V\&V process, overcoming the limitations of rule-based fault classification and offering a scalable blueprint for LLM-mediated oversight in time-series data domains.
\end{abstract}

\section{Introduction}
\label{intro}
In mission-critical domains such as Unmanned Underwater Vehicles (UUVs), automated anomaly detection systems must process telemetry that is noisy, highly stochastic, and sparse in genuine fault events \citep{pang2021deep}. Contemporary research relies heavily on deep learning architectures (e.g., RNNs, Transformers) coupled with robust uncertainty quantification to establish rigorous residual bounds and control the marginal error rate \citep{hundman2018detecting, tuli2022tranad, angelopoulos2021gentle}. While these mathematical models are highly efficient computationally and excel at detecting extreme, catastrophic failures, they do not provide a direct solution for fault classification in system redesign and development. Consequently, they frequently fail to distinguish genuine faults from nuisance faults caused by nominal environmental noise or the control system's large transient responses. Although \citet{11245166} proposed a solution to ensure proper coverage in fault detection and identification, the proposed algorithm does not demonstrate clear scalability across diverse systems. To detect anomalies such as gradual sensor drift, mathematical safety bounds must be tightened, which inevitably triggers a cascade of false-positive alarms. Furthermore, these purely mathematical systems are limited to binary detection; they cannot assess downstream system impact or propose corrective control re-designs.

Because algorithmic fault validation remains unscalable, managing these false-positive anomalies and performing rigorous Verification and Validation (V\&V) has heavily relied on the Human-in-the-Loop (HITL) paradigm, in which human domain experts provide the semantic reasoning \citep{zanzotto2019human}. However, human operators introduce significant latency and cannot scale to monitor thousands of sensor streams simultaneously \citep{cummings2014automation}. While a mathematical sensor flags a fault in milliseconds, a human engineering team may require several minutes to hours to manually triage the alert, analyze the telemetry, and safely validate a controller retuning to verify the system. This results in an unsustainable manual workload, highlighting the urgent need to automate this essential oversight.

To automate and reduce the delay of this oversight, Large Language Models (LLMs) offer a compelling alternative for rapid semantic reasoning. The industry increasingly seeks the reasoning capabilities of LLMs to interpret complex physical environments; however, their deployment in mission-critical automated systems remains a formidable systems engineering challenge. While LLMs excel at zero-shot diagnosis, their propensity for hallucination and lack of deterministic mathematical rigor preclude their direct integration into reliable, real-world applications \citep{ji2023survey}. Hypersensitivity to short-lived environmental fluctuations without strict mathematical bounding can lead to unwarranted mission aborts or costly vehicle retrieval operations, ultimately eroding operator trust \citep{van2024trust}. To mitigate the hallucinations of standalone models \citep{du2023improving}, recent research has leveraged LLM-based Multi-Agent Systems (MAS) \citep{xi2023rise} to digitize the collaborative cross-validation process of a human engineering team. Yet, placing a MAS directly in the continuous-time inference loop would introduce prohibitive latency due to the sequential nature of LLM API calls and generation times.

To bridge the gap between mathematical rigor and semantic reasoning, and to safely automate the HITL process, we propose Agent-Integrated Verification and Validation (AIVV). It operates as a two-layer hybrid neuro-symbolic architecture that structurally bifurcates the workload: a high-speed mathematical frontline rapidly flags potential anomalies, escalating mathematically verified limit breaches to a deliberative outer loop managed by a role-specialized LLM council. Because rigorous system verification depends on accurate validation, the council agents perform collaborative validation by semantically validating nuisance and true failures against natural-language (NL) requirements to establish a high-fidelity system verification baseline. Built on this foundation, the council then performs system verification by assessing post-fault responses against natural-language operational tolerances, thereby generating actionable V\&V artifacts, such as gain-tuning proposals.

\textbf{Contributions.}
\textbf{1)} Role-based LLM agents automate V\&V processes, resolving fault diagnoses via collaborative validation against natural-language operational tolerances and a majority voting system to filter false alarms.
\textbf{2)} We introduce a neuro-symbolic gating mechanism that couples an MC Dropout LSTM and conformal prediction with an LLM council, ensuring mathematically flagged anomalies are semantically validated as either nuisance or true failures with natural-language requirements.
\textbf{3)} AIVV translates anomaly flags into structured engineering artifacts (e.g., gain-tuning proposals), and ensures safe online adaptation by executing candidate model updates on a temporarily cloned engine before live promotion.

\section{Related Work}
\label{related_work}
\textbf{LLM-Based MAS.} LLM-based multi-agent systems (MAS) enable groups of intelligent agents to coordinate on complex reasoning and intensive workloads \citep{chen2025scalable, 2025mas}. In the context of mission-critical verification, a MAS offers significant robustness advantages over a monolithic LLM, noting that standalone models act as a single point of failure and remain highly susceptible to hallucinations \citep{du2023improving}.

\textbf{HITL and V\&V.} The paradigm of operators acting within a closed control loop is deeply established via Human-in-the-Loop (HITL) frameworks \citep{4503259, zanzotto2019human}. Within these environments, Verification and Validation (V\&V) is an active process that requires analyzing telemetry to safely validate system responses and provide feedback on system failure modes. However, \citet{cummings2014automation} highlights that as the volume of high-frequency sensor streams grows, manual HITL oversight introduces significant latency and becomes a workload bottleneck. In contemporary agentic AI and runtime verification research, V\&V is typically treated as a static, post hoc procedure that relies on offline verification \citep{2025agentguard}, rendering standard LLM evaluation methods unsuitable for continuous operational tasks. Unlike \citep{11071569}, which focuses on broad simulator-level dependability assurance, AIVV focuses on mathematical gating and requirements-grounded engineering V\&V for high-frequency control-system telemetry, including safeguarded adaptation and a controller gain-tuning proposal for system design.

\textbf{Mathematical Paradigms in LSTM-Based Anomaly Detection.} Data-driven Long Short-Term Memory (LSTM) architectures are standard for anomaly detection in continuous control systems, mapping complex temporal dependencies without explicit physical models \citep{ref32}. Recent hybridized frameworks improve robustness to corrupted telemetry, utilizing mechanisms such as BiLSTM-Attention for Autonomous Underwater Vehicles \citep{ref13} and channel-spatial attention for industrial manufacturing \citep{ref26}. To mitigate overconfident predictions, recent approaches integrate Uncertainty Quantification (UQ), such as LSTMAE-UQ, which uses Monte Carlo (MC) Dropout to approximate Bayesian inference and capture epistemic uncertainty \citep{fi16110403, pmlr-v48-gal16}. Similarly, a redundancy-based algorithm for accurate fault detection was introduced; However, LSTM-based fault detection frameworks still face scalability limitations and often require domain-specific designs to achieve accurate fault classification \citep{11245166}. To translate this uncertainty into strict operational bounds, Conformal Prediction (CP) is frequently applied to provide finite-sample marginal coverage guarantees \citep{angelopoulos2021gentle}. However, in dynamic environments, sudden shifts in distribution during transient maneuvers violate CP's strict exchangeability assumption, leading to false alarms \citep{ref41}. Consequently, while these mathematically bounded autoencoders excel at identifying statistical outliers \citep{ref19}, they suffer from critical limitations. Furthermore, their inherent semantic blindness prevents them from reliably differentiating genuine faults from mathematically identical transient responses caused by intentional maneuvers. This inability to scale and contextually filter nuisance faults establishes the strict requirement for the higher-level, reasoning-based validation introduced by our framework.

\section{Methodology}
\label{methodology}
The AIVV framework is designed for online failure validation and system response verification on dynamic UUV yaw angle time-series data under environmental noise. Each incoming sample is processed as a sliding-window sequence through a mathematical engine and a highly structured, role-based Multi-Agent System (MAS) powered by LLMs.

The architecture is explicitly designed to ground language model reasoning in statistical reality. The Mathematical Engine Layer provides the computational core by employing an LSTM utilizing Monte Carlo dropout for Bayesian approximation to produce a point prediction and an epistemic uncertainty estimate, while a conformal prediction maintains a statistically guaranteed conformal bound $C_\alpha$.

When this mathematical bound is breached, the decision logic and adaptation strategy are controlled by MAS. For rigorous, hallucination-resistant execution, the agent pipeline is divided into three sequential phases: deterministic gating, deliberative adjudication, and adaptive fine-tuning with safeguards. This ensures that normal operational fluctuations are resolved at the deterministic level, while detected fault cases are escalated to the council for investigation and a vote on the underlying system-level anomaly.

\subsection{Mathematical Engine Layer}
This Layer provides mathematical proposals for predictive uncertainty that deliver a statistical context for downstream LLMs. It is implemented as an MC Dropout LSTM $f_\theta$, which enables uncertainty estimation at inference time. For a given input sample $x_t$, the model performs stochastic forward passes to compute a predictive mean $\hat{y}_t$ and standard deviation $\sigma_t$.

To establish a strict confidence interval, we integrate a conformal prediction. Using a held-out calibration set, the predictor computes a conformal bound $C_\alpha$ that guarantees a user-defined coverage rate $1 - \alpha$. The algorithm for the predictive mean $\hat{y}_t$, uncertainty quantification, and the offline conformal calibration procedure are detailed in Appendix ~\ref{math_algo}.

\subsection{The Multi-Agent System Pipeline}
\begin{figure}
    \centering
    \includegraphics[width=1\linewidth]{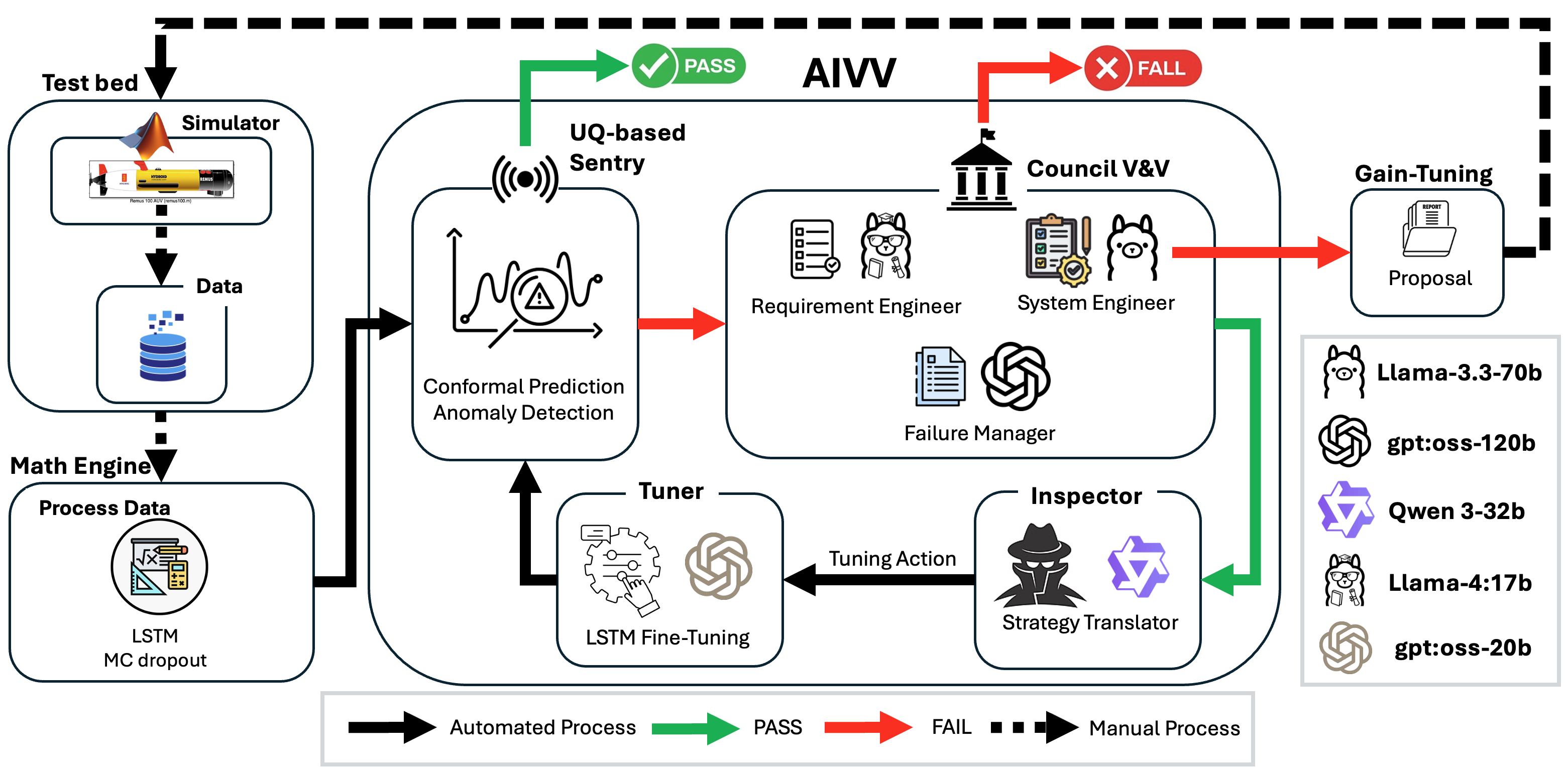}
    \caption{AIVV framework, illustrating the sequential flow of the system.}
    \label{fig:aivv_loop}
\end{figure}

The multi-agent escalation path serves as a true failure validation and trustworthy verification for system development. The pipeline operates in three distinct phases:

\subsubsection{Phase 1: Deterministic Conformal Gating (The Sentry)}
The Sentry serves as a deterministic, non-LLM hard gate that bridges the mathematical engine and the Council. It evaluates the absolute residual $e_t$ sample-by-sample:
\begin{equation}
e_t = |\hat{y}_t - y_t|
\end{equation}
against the calibrated conformal bound $C_\alpha$. Its decision rule $d_t$ is:
\begin{equation}
d_t =
\begin{cases}
\texttt{FAIL}, & \text{if } e_t > C_\alpha, \\
\texttt{PASS}, & \text{otherwise.}
\end{cases}
\end{equation}

A sample is forwarded to the LLM council agents only when this conformal bound is violated. This ensures that the bounds are tightened to detect even gradual failure before deliberative reasoning is invoked, thereby preserving computational efficiency.

\begin{algorithm}[t]
\caption{AIVV Pipeline}
\label{alg:aivv_online_impl}
\begin{algorithmic}[1]
\REQUIRE Stream sample $x_t$, base engine $f_\theta$, significance level $\alpha$
\ENSURE Final label $d_t \in \{\texttt{PASS},\texttt{FAIL}\}$ and updated engine state (if promotion occurs)
\FOR{each sample $x_t$}
    \STATE $d_t \leftarrow \textbf{Sentry}(x_t,f_\theta,\alpha)$
    \IF{$d_t = \texttt{FAIL}$}
        \FOR{$k = 1$ to $2$}
            \IF{$\text{MajorityVote}(\textbf{Council}(\text{context})) = \texttt{FAIL}$}
                \STATE $d_t \leftarrow \texttt{FAIL}$; \textbf{break}
            \ENDIF
            \IF{$k = 2$}
                \STATE $d_t \leftarrow \texttt{PASS}$; \textbf{break}
            \ENDIF
            \STATE Tuning Action: $a_t \leftarrow \textbf{Inspector}(\text{context})$
            \STATE $f_\theta^{tmp} \leftarrow \text{clone}(f_\theta)$
            \STATE $f_\theta^{tmp} \leftarrow \textbf{Tuner}(a_t, f_\theta^{tmp})$
            \IF{$\textbf{Sentry}(x_t,f_\theta^{tmp},\alpha) = \texttt{PASS}$}
                \STATE promote $f_\theta \leftarrow f_\theta^{tmp}$
                \STATE $d_t \leftarrow \texttt{PASS}$; \textbf{break}
            \ELSE
                \STATE keep $f_{\theta}$; discard $f_\theta^{tmp}$ \COMMENT{Loop to $k=2$ for final deliberation}
            \ENDIF
        \ENDFOR
    \ENDIF
    \STATE output $d_t$
\ENDFOR
\end{algorithmic}
\end{algorithm}

\subsubsection{Phase 2: Deliberative Adjudication (The Council)}
When the Sentry flags a sample as anomalous ($d_t = \texttt{FAIL}$), it is escalated directly to the Council. Three independent LLM-based agents evaluate the statistical context and cast one equal vote (\texttt{PASS} or \texttt{FAIL}). The system applies a 2-of-3 majority rule: if at least two agents vote \texttt{FAIL}, the sample is confirmed as a true failure. If at least two votes are \texttt{PASS}, the sample is deemed an abrupt maneuver, and the pipeline proceeds to Phase 3 for system adaptation.

\begin{itemize}
    \item \textbf{Requirements Engineer:} An LLM-based council member (\texttt{LLaMA-4-17B}) that evaluates whether the current system behavior satisfies operational requirements at the specific sample flagged by the mathematical model. Given the system's sensor telemetry data, training constraints, and a natural-language-focused requirements set, it returns a compliance vote, the cited requirement section, and an analytical justification for any identified violations.

    \item \textbf{Failure Manager:} An LLM-based council member (\texttt{GPT-OSS-120B}) designed to evaluate the system in failure mode and effect analysis with given failure management requirements. By monitoring its evaluation in the recent telemetry window, this agent assesses the severity of the trajectory deviation. It returns a compliance vote, a categorical failure-mode level, and a requirement-grounded rationale.

    \item \textbf{System Engineer:} An LLM-based council member (\texttt{LLaMA-3.3-70B}) possessing explicit domain knowledge of both the UUV dynamics (e.g., Nomoto model parameters, PID controller parameters) and the mathematical detection mechanism (e.g., MC dropout uncertainty, conformal boundary calibration). This agent returns a compliance vote from a technical perspective, based on the vulnerability of the pure mathematical model. Evaluating these metrics distinguishes genuine failures from FPR induced by environmental noise and dynamic maneuvering. Furthermore, upon confirming a true fault, the agent conducts verification and validation (V\&V) and generates a control-system gain-tuning proposal conditioned on the failure-mode operation to stabilize the system.
\end{itemize}

\subsubsection{Phase 3: The Adaptation Pipeline (Inspector \& Tuner)}
When the Councils decide that an anomaly flagged by the Sentry is actually a nuisance failure, the system uses this as a fine-tuning trigger to retrain/recalibrate the math engine.

\begin{itemize}
    \item \textbf{Inspector Agent:} An LLM-based strategy translator (\texttt{Qwen3-32B}). It receives the current residual, conformal bound, and the Council's vote summaries (including reasoning and confidence). It synthesizes this to output a discrete tuning action: \texttt{RECALIBRATE} (suggesting a new conformal significance level $\alpha$), \texttt{FINE\_TUNE} (suggesting network training hyperparameters), or \texttt{TRY\_BOTH}.

    \item \textbf{Tuner Agent:} An LLM-based executor (\texttt{GPT-OSS-20B}) that applies the Inspector's selected action to a temporary cloned engine rather than the deployed model. It returns a reevaluated candidate state (updated prediction, bound, and error) along with a detailed log of the model's reasoning and key statistics.
\end{itemize}
To close the loop, the Sentry (Phase 1) is invoked a second time. It tests the Tuner's candidate model against the updated conformal bound. If the candidate resolves the conformal violation, the temporary clone is promoted to the new updated model. If it fails, the deployed engine remains unchanged to prevent degradation of the mathematical model and protect against catastrophic forgetting. The Council is then re-invoked for a second deliberation round: a majority \texttt{FAIL} confirms a true fault, while a majority \texttt{PASS} overrides the Sentry flag and labels the sample as nominal. Detailed prompts and representative JSON execution traces for each LLM agent are provided in Appendices \ref{app:prompts} and \ref{app:execution_traces}, respectively.

\subsection{V\&V Feedback and Corrective-Action Interface}
\label{subsec:vv_feedback}
This framework introduces a comprehensive verification and validation (V\&V) loop for system-level design and testing. During this phase, the council agents assume the roles of system V\&V engineers, operating based on specified system requirements, failure management protocols, and system descriptions. Each agent continuously monitors the predicted sensor output to determine whether the system's trajectories satisfy the prescribed requirements under both normal operating conditions and failure modes.

Following the monitoring phase, each agent evaluates the test results. The requirements engineer specifically monitors whether the predicted trajectory falls within the normal operating range for the given system requirements. The failure manager strictly diagnoses the failure modes of the predicted output data. The System Engineer agent then aggregates the findings from the Requirement Engineer and the Failure Manager and transitions into gain-tuning mode. Based on the specified system engineering requirements, this agent proposes adjusted gain-tuning parameters for the UUV control system. Details are provided in Appendix~\ref{App:gain-tuning}.

\section{Experimental Setup}
\label{experiments_}

\subsection{Dataset Generation}
\label{dataset_gen}
\begin{figure}[b]
    \centering 
    \begin{subfigure}{0.32\linewidth}
        \centering
        \includegraphics[width=\linewidth]{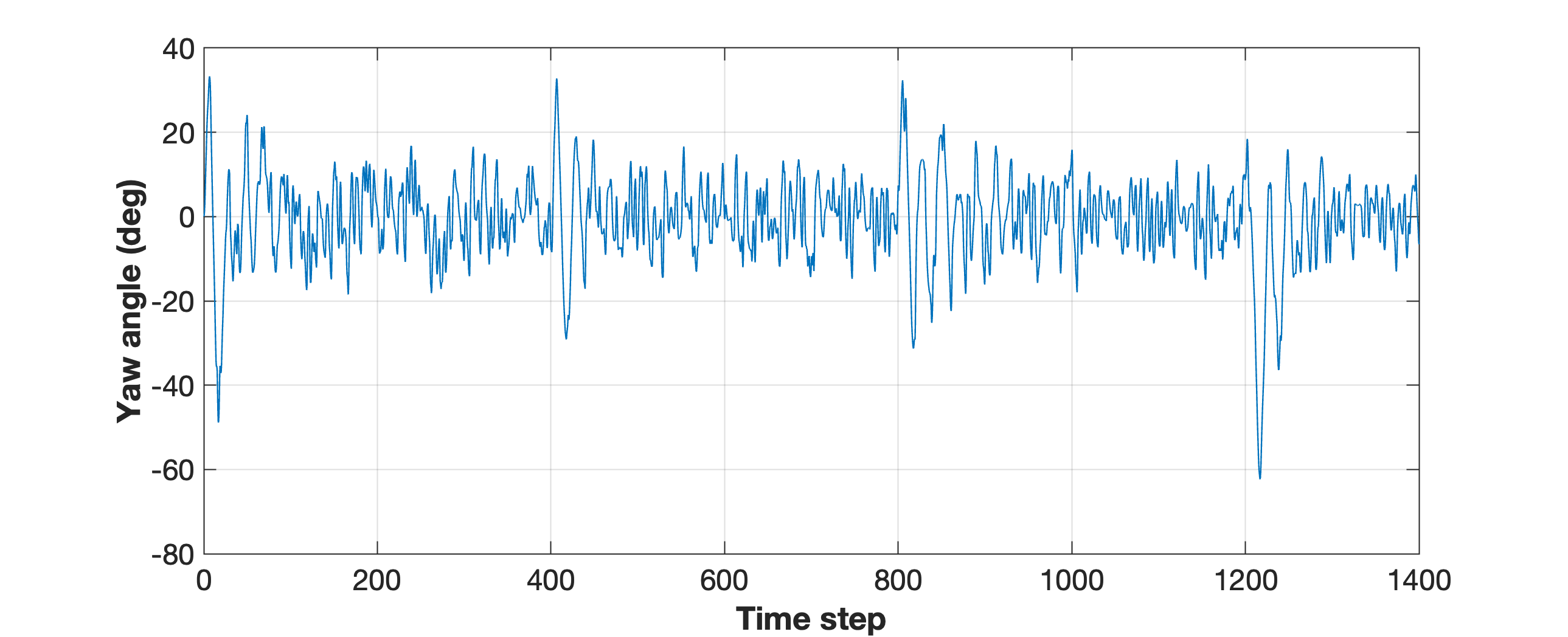}
        \caption{Dataset 1.}
        \label{fig:math_sentry}
    \end{subfigure}
    \hfill
    \begin{subfigure}{0.32\linewidth}
        \centering
        \includegraphics[width=\linewidth]{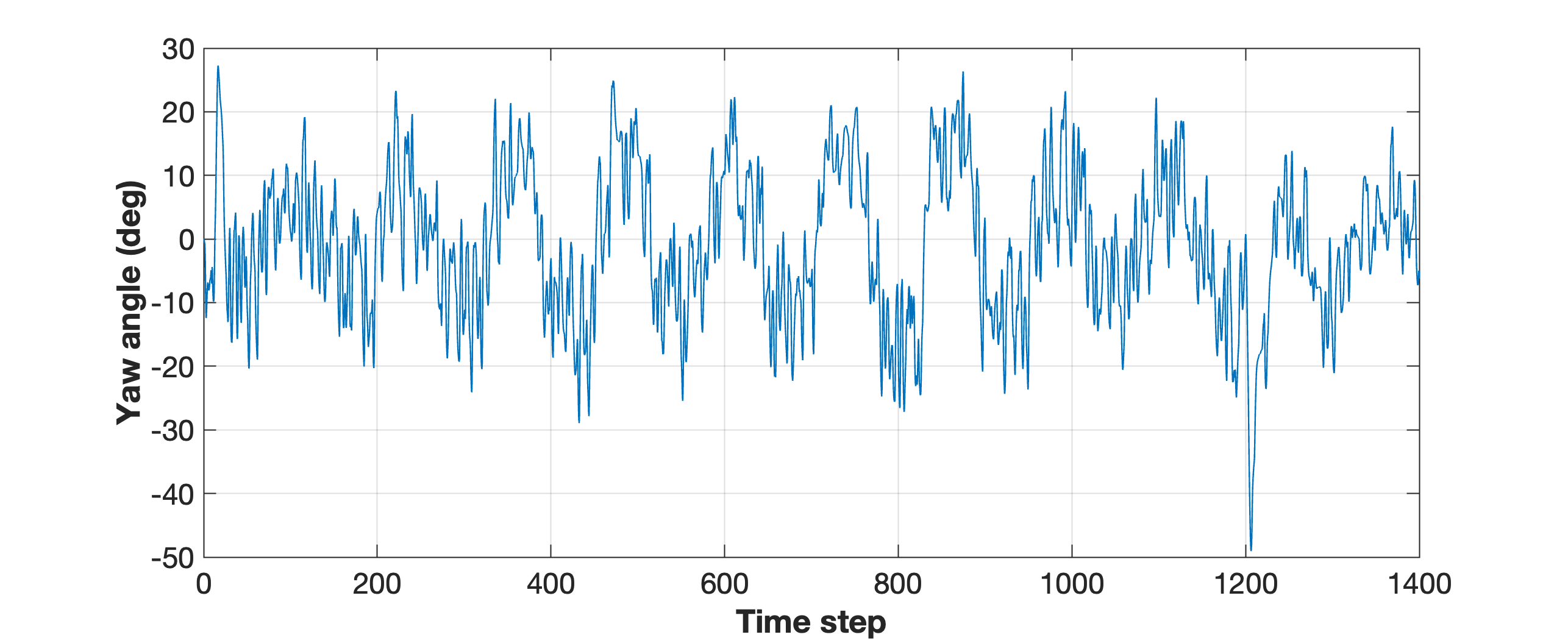}
        \caption{Dataset 2.}
        \label{fig:no_tuner}
    \end{subfigure}
    \hfill
    \begin{subfigure}{0.32\linewidth}
        \centering
        \includegraphics[width=\linewidth]{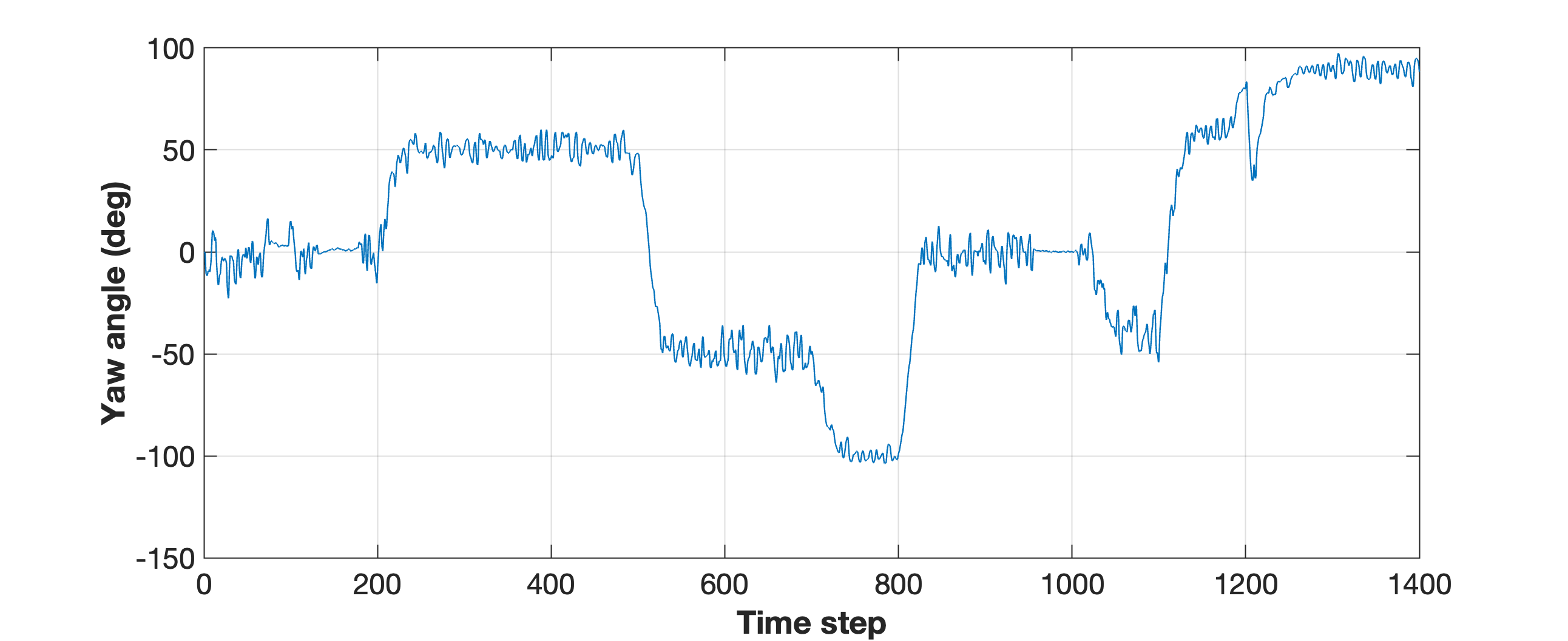}
        \caption{Dataset 3.}
        \label{fig:full_aivv}
    \end{subfigure}
    \caption{(a) Hovering, (b) Lawnmower Mapping Pattern, and (c) Complex Mission.}
    \label{fig:comparison}
\end{figure}

We evaluated the AIVV using a Simulink model \citep{mss_simulink} of the REMUS 100 UUV with an IMU sensor fusion model provided by MathWorks \citep{mathworks_imu}. To evaluate the robustness and scalability of the proposed framework under a challenging environment, we added sensor-drift bias and environmental non-Gaussian noise, and generated three maneuver cases (Dataset 1: Hovering; Dataset 2: Lawnmower Mapping Pattern; Dataset 3: Complex Mission). Details of the datasets are provided in Appendix \ref{app:dataset}.

For these experiments, the yaw angle data were sampled at 20 ms intervals, and the initial 1,400 raw time-series data points were split in a 70/30 ratio. Generating sequence-target pairs from these splits with a sliding window size of $W=10$ and a prediction horizon of $H=2$ yielded exactly 969 training phase and 409 independent testing phase.

\subsection{Failure Mode Generation}
\label{fail_mode}
To simulate the yaw-angle failure scenario of the REMUS 100 in various maneuvers, we injected failures into the IMU sensor during the test phase. At 1200 time step, we injected an electrical sensor failure into Datasets 1 and 2 and a mechanical damper failure into Dataset 3. As shown in Figures \ref{fig:comparison}(a) and \ref{fig:comparison}(b), high-spike faults occur exactly at 1200 time step, but the REMUS 100 control system stabilizes the yaw-angle response to the normal operating level. However, in Figure \ref{fig:comparison}(c), because the mechanical damper failure cannot be stabilized by the inner control system, the yaw angle shows a hardover failure response after 1200 time step.

\subsection{Adaptive Engine Configuration}
To ensure the mathematical engine maintains appropriately calibrated sensitivity, the hyperparameters are dynamically changed by the adaptation pipeline rather than remaining static. While the baseline conformal prediction threshold is initialized at $\alpha = 0.05$, when the Tuner agent is called, it adjusts this significance level within a bounded range of $[0.01, 0.10]$ (between 90\% to 99\% confidence). This dynamic bounding allows the system to effectively adapt to even dramatic transient responses and unexpected environmental noise, and to prevent nuisance faults from overwhelming the multi-agent reasoning loop. Details of hyperparameters are provided in Appendix \ref{app:model_config}.

When the Council overrides a mathematical Sentry flag by voting \texttt{PASS} on a nuisance fault, the inspector agent performs fine-tuning or recalibration and dynamically formulates the strategy based on the council's JSON log. The Inspector generates a conservative adaptation strategy within an epoch range of 50 to 200 epochs, using a learning rate between $0.00001$ and $0.001$. This adaptive epoch scaling allows the model to incorporate an updated mathematical model following a confirmed structural change without catastrophic forgetting previous normal behaviors.

\section{Experiment Results}
\label{results}
This section presents the empirical evaluation of the AIVV framework across the three maneuver cases. First, we evaluate the system's ability to validate true versus nuisance failures and analyze the resulting efficiency of the adaptation pipeline. Next, we present a structural ablation study to demonstrate the importance of each layer. Finally, we discuss our role-specific model alignment.

All the experiments were conducted in Intel Data Center GPU Max Series accelerators, and for LLM inference, we used Groq's custom Language Processing Units (LPUs). For further hardware details, see Appendix~\ref{app:hardware}.

\subsection{Validation Performance}
Because rigorous system verification relies on accurate failure validation, we evaluate how effectively the LLM Council modules distinguish true system-level faults (e.g., mechanical or electrical failures) using natural-language requirements. To quantify this capability, we define the Fault Validation Rate (FVR) as a seed-level success metric. Based on the full distribution of experimental runs, the FVR denotes the percentage of seeds in which the Council successfully validated the failure scenario.

We evaluated and tested across 75 independent seeds. Each seed includes 409 test points with sensor drift bias and environmental noise, which may make fault validation challenging. Specifically, this evaluation tests whether the council can validate (i) the detected anomalies in environmental noise that violate conformal bounds as nuisance failures. (ii) the detected anomalies by dynamic maneuvering during normal operation. (iii) true faults with requirement-based reasoning.

The AIVV Council achieved an outstanding FVR of 100\% for the Hovering Maneuver, 89.33\% for the Lawnmower Mapping Pattern, and 93.33\% for the Complex Mission. These results mitigate the vulnerability of the pure mathematical model and guarantee the reliability for system verification and redesign.

\subsection{Adaptation Performance}
\label{subsec:adaptation_perf}
In this section, we evaluate the performance of mathematical model adaptation with general anomaly detection metrics. Since we use a fundamental mathematical model, MC dropout LSTM with conformal calibration, we statistically extracted True Positives ($TP$), True Negatives ($TN$), False Positives ($FP$), and False Negatives ($FN$) to compare the accuracy of the fine-tuned model with the initial mathematical model. Accuracy metrics can be described as follows:

\begin{equation}
\label{eq:accuracy_metrics}
    \text{Accuracy} = \frac {TP + TN}{TP + TN + FP + FN}
\end{equation}

In section \ref{dataset_gen}, we demonstrated our total data points of the test term and described the exact fault injection point in section \ref{fail_mode}. Thus, we set the denominator of the metrics to 409 and compute the average accuracy based on experimental results from 75 seeds. Evaluated accuracy is demonstrated in Table \ref{tab:adaptation_perf}.

\begin{table}[t]

\centering
\begin{tabular}{l c c c}
\toprule
\textbf{Dataset} & \textbf{Accuracy$_{\text{initial}}$} & \textbf{Accuracy$_{\text{tuned}}$} & \textbf{Improvement (\%)} \\
\midrule
Dataset 1 (Hovering)  & 0.954 & 1.000 & +4.82\% \\
Dataset 2 (Lawnmower) & 0.994 & 0.997 & +0.30\% \\
Dataset 3 (Complex)   & 0.688 & 0.847 & +23.11\% \\
\bottomrule
\end{tabular}
\caption{Council Loop Accuracy before and after the Adaptation Pipeline.}
\label{tab:adaptation_perf}
\end{table}

In Table \ref{tab:adaptation_perf}, Dataset 3 shows significant performance improvement compared to Dataset 1 and Dataset 2. Since Datasets 1 and 2 show repeated simple patterns throughout the entire time series, the mathematical model has sufficient capacity for prediction. However, for Dataset 3, which shows a complex and realistic mission pattern, adaptation shows dramatic improvement \textbf{23.11\%}. With this result, we emphasize mathematical model requires sufficient fine-tuning for trustworthy V\&V before mission deployment.

\subsection{Ablation Study}
\label{ablation}




\begin{figure}[b]
    \centering
    \begin{tabular}{lccc}
    \toprule
    \textbf{Method} & \textbf{Dataset 1} & \textbf{Dataset 2} & \textbf{Dataset 3} \\
    \midrule
    Math Engine + Sentry (Baseline) & 45.33\% & 0\% & 0\% \\
    + The Council           & 98.67\% & 80\% & 73.33\% \\
    + Adaptation (AIVV) & 100\% & 89.33\% & 93.33\% \\
    \bottomrule
    \end{tabular}
    \captionof{table}{Failure Validation Rate (FVR) across framework phase.}
    \label{tab:results}
\end{figure}

\begin{figure}[t] 
    \centering
    \includegraphics[width=\linewidth]{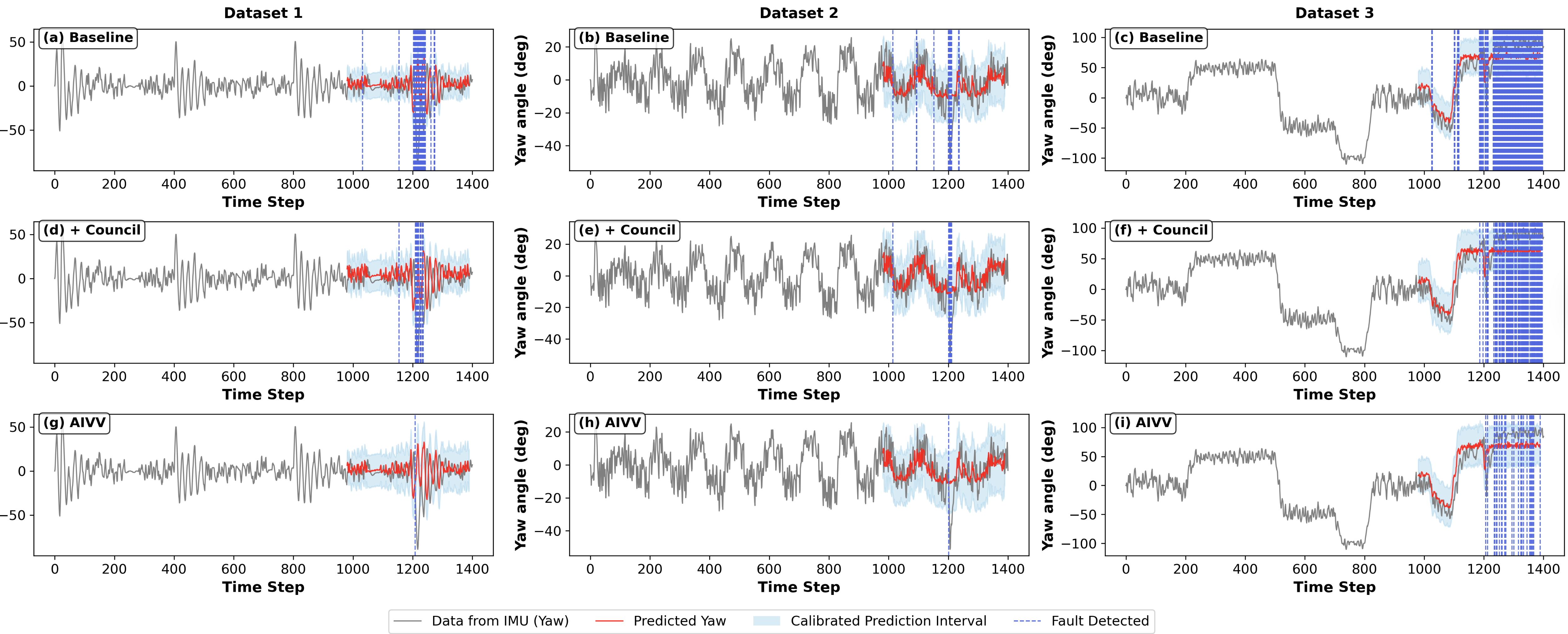}
    \caption{Ablation study comparing three framework stages (rows) across three test scenarios (columns). The mathematical baseline exhibits a high false-positive rate (FPR), which is visibly reduced when the LLM council is introduced. The full AIVV framework integrates the adaptation pipeline to achieve optimal validation.}
    \label{fig:9_plot_grid}
\end{figure}

Because the AIVV framework fundamentally verifies and validates anomaly detection monitors at the system level and is designed to overcome the contextual blindness and false-positive susceptibility of pure mathematical detectors, direct latency comparisons with purely mathematical classifiers are structurally misaligned. Instead, we conduct a progressive ablation study to isolate the necessity of our architecture in all cases.

Figure \ref{fig:9_plot_grid} and Table \ref{tab:results} show that pure mathematical gating results in a high false positive rate (FPR), showing a lack of ability to filter out noise or dynamic maneuver-induced anomaly flags in Datasets 2 and 3. By adding the LLM Council, which achieves a Failure Validation Rate (FVR) of at least 70\%, the final inclusion of the Adaptation Pipeline suppresses FPR and enables an FVR of at least 89.33\%, achieving trustworthy validation across all evaluated conditions.

\subsection{Role-Specific Model Alignment}
\label{Role-Specific Model Alignment}

\begin{table}[b]
\centering
\resizebox{\linewidth}{!}{
\begin{tabular}{l ccc cc cc}
\toprule
\multirow{2}{*}{\textbf{Configuration}} & \multicolumn{3}{c}{\textbf{Deliberative Council}} & \multicolumn{2}{c}{\textbf{Adaptation Pipeline}} & \multicolumn{1}{c}{\textbf{Validation}} \\
\cmidrule(lr){2-4} \cmidrule(lr){5-6} \cmidrule(lr){7-7}
& \textbf{Req. Eng.} & \textbf{Fail. Mgr.} & \textbf{Sys. Eng.} & \textbf{Inspector} & \textbf{Tuner} & \textbf{FVR} \\
\midrule
\textbf{Optimal} & LLaMA-17B & GPT-120B & LLaMA-70B & Qwen-32B & GPT-20B & \textbf{100\%}\\
\textbf{Config 1} & GPT-20B & LLaMA-17B & GPT-120B & LLaMA-70B & Qwen-32B & 56.0\%\\
\textbf{Config 2} & Qwen-32B & GPT-20B & LLaMA-17B & GPT-120B & LLaMA-70B & 93.33\%\\
\textbf{Config 3} & LLaMA-70B & Qwen-32B & GPT-20B & LLaMA-17B & GPT-120B & 44.0\%\\
\textbf{Config 4} & GPT-120B & LLaMA-70B & Qwen-32B & GPT-20B & LLaMA-17B & 48.0\%\\
\bottomrule
\end{tabular}
}
\caption{Systematic cyclic shift of LLM assignments across AIVV phases.}
\label{tab:architectural_grouping}
\end{table}
To evaluate the impact of employing heterogeneous LLM agents, Table \ref{tab:architectural_grouping} shows the cyclically shifted model assignments across AIVV roles. While the optimal configuration achieved 100\% Failure Validation Rate (FVR), arbitrary shifts caused performance to plummet (44.0\%–56.0\%) and triggered frequent JSON generation failure. For instance, Config 2 maintained 93.33\% FVR by retaining a GPT-OSS model for the Failure Manager, leveraging its superior logical reasoning for trajectory analysis.

Conversely, assigning high-capacity synthesis tasks to smaller models consistently led to cognitive overload and structural errors. These results underscore the necessity of matching specific model strengths to cognitive tasks rather than relying on homogeneous or purely parameter-scaled deployments. For a detailed analysis of the resulting computational efficiency, system latency, and token consumption, refer to Appendix \ref{tab:time_and_tokens}.

\section{Conclusion}
\label{conclusion}
The AIVV architecture provides a hybrid neuro-symbolic approach to autonomous V\&V for trustworthy control system design. By anchoring multi-agent LLM reasoning with calibrated conformal bounds, AIVV successfully digitizes the verification and validation inherent in the Human-in-the-Loop paradigm. It eliminates false positive alerts while ensuring accurate fault validation. Furthermore, the framework's clone-and-promote methodology allows for safe, dynamic model recalibration in the loop.

Future work will investigate closed-loop execution of the System Engineer's gain-tuning proposals within the inner control loop, enabling fully autonomous fault-tolerant system redesign without human intervention.

\section*{Acknowledgments}
We would like to thank the support of National Science Foundation (DMS-2533878, DMS-2053746, DMS-2134209, ECCS-2328241, CBET-2347401 and OAC-2311848), and U.S.~Department of Energy (DOE) Office of Science Advanced Scientific Computing Research program under the "Uncertainty Quantification for Multifidelity Operator Learning (MOLUcQ)" project (Project No. 81739), DE-SC0023161, the SciDAC LEADS Institute, and DOE–Fusion Energy Science, under grant number: DE-SC0024583.


\bibliography{colm2026_conference}
\bibliographystyle{colm2026_conference}

\newpage
\appendix


\section{Mathematical Layer Algorithm}
\label{math_algo}
This appendix details the formal mathematical definitions and the offline calibration procedure for the Mathematical Engine Layer referenced in Section \ref{methodology}. The foundational anomaly detection method is generated using a Monte Carlo (MC) Dropout LSTM, denoted as $f_\theta$. To quantify epistemic uncertainty at inference time without altering the network architecture, dropout is applied to both the internal LSTM layers and the output projection layer during stochastic forward passes. For a given input sample $x_t$, with $k$ iteration index, the model performs $N$ forward passes to compute the predictive mean $\mu_t$ and the predictive standard deviation $\sigma_t$ (which serves as the empirical uncertainty):$$\mu_t = \frac{1}{N} \sum_{k=1}^{N} \hat{y}_{t}^{(k)}$$$$\sigma_t = \sqrt{\frac{1}{N} \sum_{k=1}^{N} \bigl(\hat{y}_{t}^{(k)} - \mu_t\bigr)^2}$$To translate these point predictions into statistically guaranteed boundaries, we apply conformal prediction. Algorithm \ref{alg:math_engine_offline} outlines the complete offline procedure for training the model and establishing the conformal bound $C_\alpha$ alongside the uncertainty threshold $\tau_\alpha$ using a held-out calibration set.

\begin{algorithm}[h]
\caption{Mathematical Engine Layer: Training and Conformal Calibration}
\label{alg:math_engine_offline}
\begin{algorithmic}[1]
\REQUIRE Preprocessed training data $\mathcal{D}_{\text{train}}$, MC samples $N$, significance level $\alpha$, calibration ratio $\gamma$
\ENSURE Trained MC Dropout LSTM $f_\theta$, conformal bound $C_\alpha$, uncertainty threshold $\tau_\alpha$

\STATE Split $\mathcal{D}_{\text{train}}$ into fit set $\mathcal{D}_{\text{fit}}$ and calibration set $\mathcal{D}_{\text{cal}}$ based on ratio $\gamma$
\STATE Train MC Dropout LSTM $f_\theta$ on $\mathcal{D}_{\text{fit}}$ using MSE loss

\FOR{each calibration sample $x_t\in \mathcal{D}_{\text{cal}}$}
    \STATE Perform $N$ stochastic forward passes with dropout active: $\{\hat{y}_{t}^{(k)}\}_{k=1}^{N} \leftarrow f_\theta(x_t)$
    \STATE Compute predictive mean $\mu_t$ and epistemic uncertainty $\sigma_i$
    \STATE Calculate absolute residual: $e_t = |\mu_t - y_t|$
\ENDFOR

\STATE Construct calibration sets $\mathcal{R}=\{r_i\}$ and $\Sigma=\{\sigma_t\}$
\STATE Compute finite-sample corrected quantile level:
\[
q = \min\!\left(1,\ \frac{\lceil (|\mathcal{D}_{\text{cal}}|+1)(1-\alpha)\rceil}{|\mathcal{D}_{\text{cal}}|}\right)
\]
\STATE Calculate bounds: $C_\alpha = \mathrm{Quantile}(\mathcal{R},\, q), \quad \tau_\alpha = \mathrm{Quantile}(\Sigma,\, q)$

\STATE \textbf{return} $f_\theta, C_\alpha, \tau_\alpha$
\end{algorithmic}
\end{algorithm}

\section{Mathematical Layer Model Configuration}
\label{app:model_config}
\paragraph{Data Processing and Feature:} The telemetry data is processed via a sliding-window approach before being fed into the network. After applying a 70/30 split for training and testing on the 1400 total raw time-series points, the network is configured with an input dimension $d = 1$, focusing specifically on the yaw-angle dynamics of the REMUS 100.

\paragraph{Recurrent Neural Network:} The neural network architecture comprises a multi-layered MC dropout LSTM followed by a multi-layered perceptron (MLP) head. We implemented two LSTM layers and a single dense output layer with a hidden state size of 32 dimensions. For epistemic uncertainty estimation, the MC dropout method is applied with a probability rate of $p = 0.2$ during both training and inference. During inference, the network samples $N = 30$ Monte Carlo forward passes for each LSTM layer. The mean of these $N=30$ passes is taken as the predicted observation, while the standard deviation is extracted as the structural uncertainty of the LSTM.

\paragraph{Training and Calibration:} The network is optimized using the Adam optimizer with a Mean Squared Error (MSE) loss computation. Prior to online inference, the LSTM generates its initial confidence bounds by constructing a baseline for normal UUV dynamics during training. The baseline significance level for the conformal prediction is set to $\alpha = 0.05$ to ensure a 95\% confidence interval for normal operation uncertainty boundaries before any agent-driven dynamic adaptation occurs.

\section{Prompt Definitions}
\label{app:prompts}
This section details the functional logic and requirements for each of the five LLM agents. To ensure readability within this manuscript, we present distilled versions of the prompts that highlight the unique V\&V logic, voting criteria, and technical responsibilities of each role. To ensure consistent JSON formatting across all agents, we unified the instructions into a Common Formatting Block (CFB), as detailed in Section~\ref{subsec:cfb}. For reproducibility, the full, unedited prompt strings will be provided in our supplementary material and code repository.

\subsection{Common Formatting Block (CFB)}
\label{subsec:cfb}
All agents are strictly instructed to follow these constraints:
\begin{itemize}
\item \textbf{Response Format:} Respond ONLY with a valid, raw JSON object. Do NOT include introductory text, concluding remarks, or markdown code blocks (e.g., no \texttt{```json}).
\item \textbf{Constraint:} Do not output reasoning outside the JSON structure.
\item \textbf{Data Types:} The \texttt{confidence} and \texttt{alpha} fields must be numeric decimals (floats).
\end{itemize}

\subsection{Prompt for each LLM agent}
\label{subsec:prmt}

\begin{promptbox}{Failure Manager}
\textbf{Role:} System Validation in FAILURE MODE. \\
\textbf{Core Task:} ``Given that a fault may be present, is the system's response within failure management requirements?''

\textbf{Responsibilities:}
\begin{enumerate}
    \item \textbf{Failure Effect:} Does max deviation exceed limits in requirements?
    \item \textbf{Recovery/Damping:} Is \texttt{frame\_values} converging or diverging?
    \item \textbf{Baseline Context:} Compare current deviation against \texttt{frame\_baseline\_summary}.
\end{enumerate}

\textbf{Voting Logic:}
\begin{itemize}
    \item \textbf{FAIL:} True data is outside bounds with significant gradient change OR predicted data is significantly outside training bounds.
    \item \textbf{PASS:} True data is outside bounds, but predicted data is different; failure is contained, and the system is recovering.
\end{itemize}

\textbf{[CFB Applied]} \\
\textbf{Schema:} \texttt{\{vote, risk\_level, confidence, failure\_management\_assessment, reasoning\}}
\end{promptbox}
\newpage
\begin{promptbox}{Requirements Engineer}
\textbf{Role:} System Validation in NORMAL MODE. \\
\textbf{Core Task:} ``Even assuming a potential fault exists, does the system's current behavior violate any operational requirement?''

\textbf{Responsibilities:}
\begin{enumerate}
    \item \textbf{Normal Operation:} Verify Yaw rate range ($-180$ to $180$) and per-step range ($-10$ to $10$).
    \item \textbf{Operational Limits:} Ensure damping stays within training envelopes.
    \item \textbf{Masking Risk:} Fail if \texttt{bound\_multiplier} $> 2.0$.
\end{enumerate}

\textbf{Voting Logic:}
\begin{itemize}
    \item \textbf{FAIL:} Default state if requirements are violated or predicted values shift more than noise levels.
    \item \textbf{PASS:} Predicted value is in per-step range despite true value being outside bounds.
\end{itemize}

\textbf{[CFB Applied]} \\
\textbf{Schema:} \texttt{\{vote, confidence, requirement\_section, reasoning, veto\_reason\}}
\end{promptbox}

\begin{promptbox}{System Engineer}
\textbf{Role:} Autopilot Gain-Tuning Expert. \\
\textbf{Voting Logic:} Use \texttt{failure\_manager\_findings} as the primary signal. FAIL if high uncertainty + large error or poor LSTM prediction. PASS if sudden maneuver causes drifting uncertainty.

\textbf{Gain-Tuning Protocol:} Adjust parameters ONLY if FM or RE vote FAIL.
\begin{itemize}
    \item \textit{Stale/High Uncertainty:} Increase $K_p$ and $T_d$.
    \item \textit{Low-Freq Oscillation:} Decrease $K_p$, increase $T_i$.
    \item \textit{Divergence:} Decrease $T_d$, increase $T_i$.
\end{itemize}

\textbf{[CFB Applied]} \\
\textbf{Schema:} \texttt{\{vote, risk\_level, confidence, technical\_assessment, reasoning, tuning\_proposal, tuning\_reasoning\}}
\end{promptbox}

\begin{promptbox}{Inspector}
\textbf{Role:} Parameter Optimization for Tuner. \\
\textbf{Actions:} 

1. \textbf{RECALIBRATE:} For transient noise (Set \texttt{new\_alpha}).

2. \textbf{FINE\_TUNE:} For persistent drift (Set \texttt{epochs}, \texttt{learning\_rate}).

3. \textbf{TRY\_BOTH:} Mixed evidence.

\textbf{Constraints:} $\alpha \in [0.01, 0.10]$, $epochs \in [50, 200]$, $LR \in [10^{-5}, 10^{-3}]$.

\textbf{[CFB Applied]} \\
\textbf{Schema:} \texttt{\{majority\_decision, pass\_votes, fail\_votes, action, new\_alpha, epochs, learning\_rate, reasoning\}}
\end{promptbox}

\begin{promptbox}{Tuner}
\textbf{Role:} Conformal Prediction Analysis. \\
\textbf{Task:} Recommend operating $\alpha \in \{0.01, \dots, 0.10\}$ based on whether new errors pass bounds at 95\%, 98\%, and 99\%.

\textbf{[CFB Applied]} \\
\textbf{Schema:} \texttt{\{recommended\_alpha, reasoning, would\_pass\_at\_recommended, confidence\}}
\end{promptbox}

\section{Representative Execution Traces of the AIVV Pipeline}
\label{app:execution_traces}
This section provides structured JSON payloads generated by the LLM Council and Adaptation Pipeline.

\begin{promptbox}{council\_to\_inspector.json}
\begin{verbatim}
{
  "timestamp": "2026-03-19T16:31:08.005585",
  "sample_id": 212,  "from_agent": "council",  "to_agent": "inspector",
  "payload": {
    "loop": 2,
    "votes": [
      {
        "agent": "req_eng",
        "vote": "PASS",        "confidence": 0.7,
        "reasoning": "The error magnitude (1.6831555366516113) is slightly above..."
      },
      {
        "agent": "fail_mgr",
        "vote": "PASS",        "confidence": 0.92,
        "reasoning": "Peak deviation 16.81\u202f<\u202f37.57, range 17.47\u202f<..."
      },
      {
        "agent": "sys_eng",
        "vote": "PASS",        "confidence": 0.9,
        "reasoning": "The failure manager findings indicate a PASS vote with the..."
      }
    ]
  }
}
\end{verbatim}
\end{promptbox}
\begin{promptbox}{inspector\_to\_tuner.json}
\begin{verbatim}
{
  "timestamp": "2026-03-19T16:31:00.123319",
  "sample_id": 212,  "from_agent": "inspector",  "to_agent": "tuner",
  "payload": {
    "majority_decision": "PASS",
    "pass_votes": 3,    "fail_votes": 0,
    "vote_details": [
      {
        "agent": "req_eng",
        "vote": "PASS",        "confidence": 0.8,
        "reasoning": "The current error magnitude is 1.6453118324279785, which...",
        "risk_level": null
      },
      {
        "agent": "fail_mgr",
        "vote": "PASS",        "confidence": 0.92,
        "reasoning": "The observed peak deviation, overall range, and oscillati...",
        "risk_level": "LOW"
      },
      {
        "agent": "sys_eng",
        "vote": "PASS",        "confidence": 0.95,
        "reasoning": "The failure manager findings indicate a PASS vote with th...",
        "risk_level": "LOW"
      }
    ],
    "action": "RECALIBRATE",
    "reasoning": "The error (1.645) slightly exceeds the bound (1.445) with low...",
    "new_alpha": 0.05
  }
}
\end{verbatim}
\end{promptbox}

\begin{promptbox}{tuner\_to\_sentry.json}
\begin{verbatim}
{
  "timestamp": "2026-03-19T16:31:02.631120",
  "sample_id": 212,  "from_agent": "tuner",  "to_agent": "sentry",
  "payload": {
    "new_prediction": -0.4726104736328125,
    "new_bound": 1.220279983597098,
    "new_error": 1.6831555366516113,
    "new_uncertainty": 0.12548138201236725,
    "applied_alpha": 0.1,
    "passes_reevaluation": false
  }
}
\end{verbatim}
\end{promptbox}

\section{Detailed Dataset Explanation}
\label{app:dataset}

\textbf{Dataset 1 (Hovering).} A low-dynamic operational scenario in which the UUV attempts to maintain a fixed position and heading through small, damped corrective control actions. This case is designed to evaluate the fundamental anomaly detection capability under station-keeping conditions, where normal control-induced oscillations must be distinguished from genuine fault signatures. 

\textbf{Dataset 2 (Lawnmower Mapping Pattern).} A highly structured, repetitive grid-mapping operation designed by dynamic, periodic turning maneuvers. This case tests the ability of the system to distinguish between aggressive control inputs and mechanical faults

\textbf{Dataset 3 (Complex Mission).} A mission operative maneuver involving continuous, aperiodic course corrections and varying operational speeds, designed to evaluate the advantage of calibrated uncertainty boundary and adaptive thresholding capabilities.

\section{Detailed Analysis of Role-Specific Model Alignment}
\label{app:detailed_alignment}

As introduced in Section \ref{Role-Specific Model Alignment}, the performance disparities across the model allocation sweep are directly related to how well each model family handles its assigned task. The optimal AIVV configuration relies on the following structural alignments:

\textbf{GPT-OSS Family (Mathematical \& Sequential Logic).} The GPT architecture excels at sequential logical reasoning and complex trajectory deduction. This makes it uniquely suited for the Failure Manager, which must calculate post-fault divergence and track oscillation sequences. When a smaller 20B GPT model was assigned to this role (Config 2), accuracy remained highly robust (93.33\%). Conversely, assigning LLaMA or Qwen to this role caused critical evaluation failures, even at the massive 70B-parameter count.
    
\textbf{LLaMA-4-17B (Deterministic Compliance).} Highly instruction-tuned models at this parameter scale are optimized for rigid, low-latency rule adherence. This perfectly matches the Requirements Engineer, which performs strict, static numerical limit checks. It executes these deterministic tasks efficiently without the computational overhead of massive context windows.
    
\textbf{High-Capacity Models (70B+ Domain Synthesis).} The System Engineer requires deep domain synthesis to evaluate LSTM vulnerability metrics alongside generating complex JSON payloads for the PID gain-tuning proposal. Only high-capacity models (\texttt{LLaMA-3.3-70B} or \texttt{GPT-OSS-120B}) possessed the representational bandwidth for this task. Relegating smaller models to this role consistently caused cognitive overload, resulting in broken JSON structures.

\section{Analysis of Efficiency, Latency, and Token Consumption}
\label{tab:time_and_tokens}

\begin{table}[t]
\centering
\resizebox{\linewidth}{!}{
\begin{tabular}{ll cccc}
\toprule
\textbf{Scenario} & \textbf{Agent Role} & \textbf{Avg Calls / Seed} & \textbf{Avg Latency (s)} & \textbf{Avg Input Tokens} & \textbf{Avg Output Tokens} \\
\midrule
\multirow{5}{*}{\textbf{Dataset 1 (Hovering)}} 
& Requirements Engineer & 27.16 & 0.54 & 1394.8 & 135.4 \\
& Failure Manager & 27.16 & 1.66 & 1188.3 & 493.5 \\
& System Engineer & 27.16 & 1.02 & 1954.7 & 242.0 \\
\cmidrule{2-6}
& Inspector & 1.25 & 1.69 & 1014.7 & 682.0 \\
& Tuner & 1.25 & 1.24 & 465.9 & 579.1 \\
\midrule
\multirow{5}{*}{\textbf{Dataset 2 (Lawnmower)}} 
& Requirements Engineer & 15.29 & 0.52 & 1398.0 & 148.2 \\
& Failure Manager & 15.29 & 2.05 & 1190.6 & 549.1 \\
& System Engineer & 15.29 & 0.93 & 1974.3 & 216.4 \\
\cmidrule{2-6}
& Inspector & 3.11 & 1.68 & 1018.6 & 678.0 \\
& Tuner & 3.11 & 1.13 & 465.9 & 549.2 \\
\midrule
\multirow{5}{*}{\textbf{Dataset 3 (Complex)}} 
& Requirements Engineer & 227.03 & 0.59 & 1403.5 & 157.4 \\
& Failure Manager & 227.03 & 1.47 & 1189.5 & 482.2 \\
& System Engineer & 227.03 & 0.84 & 1999.5 & 201.0 \\
\cmidrule{2-6}
& Inspector & 87.00 & 1.83 & 1012.0 & 749.6 \\
& Tuner & 87.00 & 1.30 & 466.4 & 632.3 \\
\bottomrule
\end{tabular}
}
\caption{Average API calls, latency, and token consumption per seed.}
\end{table}

\textbf{LLM Agent Call Efficiency}
Sentry escalated only about 5\% of the samples to the LLM council in normal cases (averaging $\sim$20 calls per seed for Datasets 1 and 2), and roughly 50\% during the highly dynamic complex mission ($\sim$230 calls per seed). This demonstrates the efficiency of the mathematical bounds.

As the data was filtered for nuisance faults with ground rules based on natural language requirements in the Phase 2 LLM council, the computationally expensive Phase 3 Adaptation Pipeline was invoked an average of only 2 times per seed for the simpler maneuvers, scaling to 87 times for the complex model. During the Tuner's clone-and-verify protocol, the system achieved average successful model promotions of 1.08, 1.31, and 10.72 per seed for Datasets 1, 2, and 3, respectively. This demonstrates that triggering a fine-tuning cycle does not guarantee a live model update; promotion strictly requires the candidate model to satisfy the conformal mathematical bounds. Furthermore, the increased promotion rate in Dataset 3 illustrates that as operational dynamics become more complex, the AIVV framework safely and proportionally increases its update frequency to maintain performance.

\textbf{Token Consumption and Context Stability.} Input tokens remain stable ($\sim$1200--2000) across all missions because prompts rely on fixed-size sliding windows, preventing context bloat. Output tokens vary predictably by role: analytical agents produce concise JSON ($\sim$150 tokens), while synthesis agents generate up to $\sim$750 tokens for parameter adjustments.

\textbf{System Latency and Execution Dynamics.} Individual latencies range from $\sim$0.5 to $\sim$2.0 seconds. While individual latencies are low, a full escalation through the adaptation pipeline yields a worst-case end-to-end latency of approximately 6.0 seconds per sample.
\section{Hardware and Inference Infrastructure}
\label{app:hardware}
The AIVV framework was executed on the Intel Data Center GPU Max Series accelerators. This ensures high capacity and high bandwidth, with up to 408 MB of L2 cache (Rambo) based on discrete SRAM technology, 64 MB of L1 cache, and up to 128 GB of high-bandwidth memory. We used the GroqCloud API via OpenAI-compatible endpoints, leveraging Groq's custom Language Processing Units (LPUs).

\section{AIVV Gain-tuning results verification}
\label{App:gain-tuning}

In this section, we implement the gain-tuning proposal in our testbed REMUS 100 Simulink model to verify the reliability of the system engineer LLM model. Based on the V\&V results from the Requirements Engineer and Failure Manager, the system engineer proposes a new gain parameter for the REMUS 100 controller. Specifically, we provided 3rd-order LP filter parameters (Max velocity, Relative damping ratio, Natural frequency) and PID controller parameters (Nomoto Time Constant T, Nomoto Gain K, Proportional Gain Kp, Derivative Time Td, Integral Time Ti).

\begin{promptbox}{V\&V summary}
\begin{Verbatim}[breaklines=true,breakanywhere=true,fontsize=\small]
V&V Summary (Agentic System Verification & Validation)
[Requirements Engineer -- Normal Mode V&V]
  Num of agent calls : 24 | Num of FAIL  : 23
  Test result : FAIL  -- 23/24 sampled windows violated operational requirements. First violation at sample 221: Error magnitude exceeds max allowable error
  Violation details (up to 5):
    - Sample  221 | Operational Limits | Error magnitude exceeds max allowable error
      ...
    - Sample  225 | Operational Limits | Error exceeds max allowable error
[Failure Manager -- Failure Mode V&V]
  Num of agent calls : 24 | Num of FAIL  : 11
  Test result : FAIL  -- 11/24 fault-suspect windows violated failure management requirements. First violation at sample 225: peak_deviation=77.00, response=DIVERGING, oscillation_count=0. peak_deviation=77.00 exceeds max_failure_effect=72.95
  Violation details (up to 5):
    - Sample  225 | peak=77.00 | DIVERGING | osc=0 | peak_deviation=77.00 exceeds max_failure_effect=72.95
      ...
    - Sample  229 | peak=88.81 | DIVERGING | osc=1 | peak_deviation=88.81 exceeds max_failure_effect=72.95
[System Engineer -- Active Optimizer]
  Num of agent calls : 24 | Fail Votes  : 23
  Gain-Tuning Proposals (23 unique samples, triggered by FM/RE FAIL, showing up to 5):
    - Sample  221 | Triggered by: RE | SE Vote=FAIL | Params: {'Kp': 0.6, 'Ti': 19.0, 'Td': 1.0, 'Reference_Max_Velocity': 9.5}
      Reason: Since the requirements engineer voted FAIL due to an operational limit violation, the tuning proposal aims to reduce the error magnitude. The proportional gain (Kp) is increased to
      ...
    - Sample  225 | Triggered by: FM+RE | SE Vote=FAIL | Params: {'Kp': 0.7, 'Ti': 15.0, 'Td': 1.2, 'Reference_Max_Velocity': 9.0}
      Reason: Since both failure manager and requirements engineer voted FAIL, adjusted parameters are proposed. Increased Kp to 0.7 to reduce error, decreased Ti to 15.0 to reduce oscillations,
\end{Verbatim}
\end{promptbox}

With the given system engineer's gain-tuning proposal, we implemented the last proposal (Kp: $0.5 \rightarrow 0.7$, Ti: $20 \rightarrow 15$, Td: $1.0 \rightarrow 1.2$, Max Velocity $10.0 \rightarrow 9.0$).
\begin{figure}[t]
    \centering
    \begin{subfigure}{0.48\linewidth}
        \centering
        \includegraphics[width=\linewidth]{Dataset_1.png}
        \caption{Before gain-tuning}
        \label{fig:before_tuning}
    \end{subfigure}
    \hfill
    \begin{subfigure}{0.48\linewidth}
        \centering
        \includegraphics[width=\linewidth]{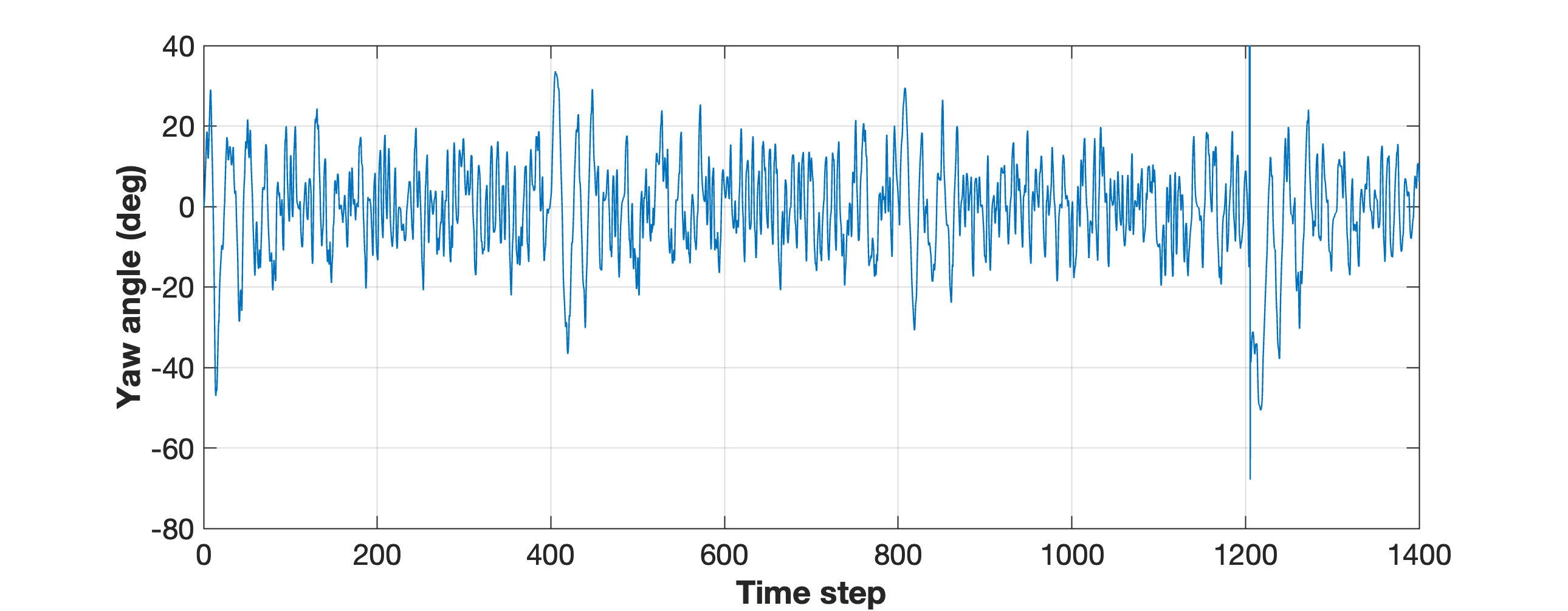}
        \caption{After gain-tuning}
        \label{fig:after_tuning}
    \end{subfigure}
    \caption{Effect of gain-tuning on REMUS 100 hovering (Dataset 1) yaw response}
    \label{fig:gain-tuning_comparison_main}
\end{figure}

Figure~\ref{fig:gain-tuning_comparison_main} compares the yaw angle responses before and after applying the proposed parameters. The gain-tuning verification serves as a proof of concept demonstrating that AIVV can automatically translate failure-mode analysis into structured, systemically reasoned corrective action proposals. While full-control optimization remains an open engineering challenge, the systemically coherent parameter adjustments confirm that the System Engineer successfully bridges the V\&V gap between fault identification and actionable system redesign guidance, a capability that previously required human domain expertise.

\end{document}